\documentclass{article}

\usepackage{microtype}
\usepackage{graphicx}
\usepackage{subfigure}
\usepackage{booktabs} % for professional tables
\usepackage{enumitem}

\usepackage{hyperref}

\usepackage{amsfonts}
\usepackage{calrsfs}
\DeclareMathAlphabet{\pazocal}{OMS}{zplm}{m}{n}
\usepackage{amsmath}

\newcommand{\eg}{{\em e.g.}}

\newcommand{\ie}{{\em i.e.}}

\newcommand{\RNum}[1]{\uppercase\expandafter{\romannumeral #1\relax}}

\usepackage[accepted]{icml2020}

\icmltitlerunning{MO-PaDGAN: Generating Diverse Designs with Multivariate Performance Enhancement}

\begin{document}
%MO-PaDGAN: Multiple Objective Performance augmented Diverse GAN

\twocolumn[
\icmltitle{MO-PaDGAN: Generating Diverse Designs with \\
Multivariate Performance Enhancement}




\begin{icmlauthorlist}
\icmlauthor{Wei Chen}{to}
\icmlauthor{Faez Ahmed}{goo}
\end{icmlauthorlist}

\icmlaffiliation{to}{Siemens Corporate Technology, Princeton, NJ 08540}
\icmlaffiliation{goo}{Northwestern University, Evanston, IL 10601}

\icmlcorrespondingauthor{Wei Chen}{chen.wei@siemens.com}
\icmlcorrespondingauthor{Faez Ahmed}{faez@northwestern.edu}

\icmlkeywords{Determinantal Point Processes, Generative Adversarial Network, Design Generation}

\vskip 0.3in
]



\printAffiliationsAndNotice{}  % leave blank if no need to mention equal contribution
% \printAffiliationsAndNotice{} % otherwise use the standard text.

\begin{abstract}% 4--6 sentences
%Existing Deep generative models
%are proven to be a 
Deep generative models have proven useful for automatic design synthesis and design space exploration. However, they face three challenges when applied to engineering design: 1)~generated designs lack diversity, 2)~it is difficult to explicitly improve all the performance measures of generated designs, and 3)~existing models generally do not generate high-performance novel designs, outside the domain of the training data. To address these challenges, we propose MO-PaDGAN, which contains a new Determinantal Point Processes based loss function for probabilistic modeling of diversity and performances. Through a real-world airfoil design example, we demonstrate that MO-PaDGAN expands the existing boundary of the design space towards high-performance regions and generates new designs with high diversity and performances exceeding training data.
%Unlike typical generative models that usually generate new designs by interpolating within the boundary of training data, we show that MO-PaDGAN expands the design space boundary outside the training data towards high-quality regions. 
%The proposed method is broadly applicable to many tasks including design space exploration, design optimization, and creative solution recommendation.
\end{abstract}

\setlength{\textfloatsep}{4pt plus 1.0pt minus 1.0pt}
\setlength{\floatsep}{4pt plus 1.0pt minus 1.0pt}
\setlength{\intextsep}{4pt plus 1.0pt minus 1.0pt}

\setlength{\belowdisplayskip}{4pt} \setlength{\belowdisplayshortskip}{5pt}
\setlength{\abovedisplayskip}{4pt} \setlength{\abovedisplayshortskip}{5pt}

% \setlength{\abovedisplayskip}{5pt}
% \setlength{\belowdisplayskip}{5pt} % for equations
% % \setlength{\abovecaptionskip}{0pt}
% \setlength{\belowcaptionskip}{0pt}
%  %\addtolength{\textfloatsep}{-30pt} % for figures
% \setlength{\textfloatsep}{2pt plus 1.0pt minus 1.0pt}
% \setlength{\abovecaptionskip}{1pt plus 3pt minus 3pt}

%%%%%%%%%%%%%%%%%%%%%%%%%%%%%%%%%%%%%%%%%%%%%%%%%%%%%%%%%%%%%%%%%%%%%%
\section{Introduction}

\begin{figure*}[!ht]
\centering
\includegraphics[width=0.9\textwidth]{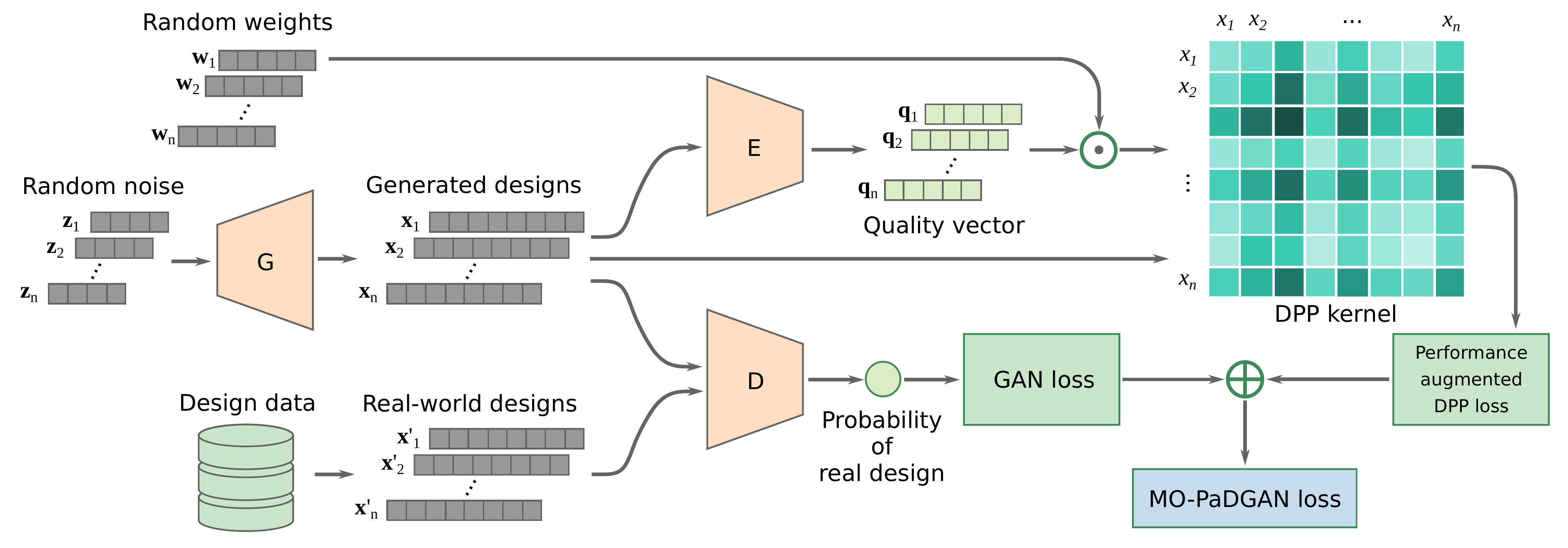}
\vspace*{-4mm}
\caption{Architecture of MO-PaDGAN. The operator $\odot$ denotes performance aggregation.}
\label{fig:architecture}
\end{figure*}

A designer wants good design solutions which are creative and meets multiple performance requirements. By manually and iteratively exploring design ideas using experience and design heuristics, the designers take the risks of 1)~wasting time on evaluating unfavorable candidates and 2)~not having sufficient width/depth for exploration/exploitation. While recent advances in machine learning assisted automatic design synthesis and design space exploration are promising, the current methods are still far from this ideal picture. To model a design space, researchers have used deep generative models like variational autoencoders (VAEs)~\cite{kingma2013auto} and generative adversarial networks (GANs)~\cite{goodfellow2014generative}, as they can learn the distribution of existing designs. The hope is that by learning an underlying low-dimensional \textit{latent space}, design exploration can be more efficient due to the reduced dimensionality~\cite{chen2017design,chen2019synthesizing,chen2019aerodynamic}. 
However, unlike image generation tasks where these generative models are commonly applied, engineering design problems typically have multiple performance measures, each of which quantifies how well a design achieves its intended goals. For example, beam design problems often have the compliance ~\cite{bendsoe_topology_2004} or both the compliance and natural-frequency~\cite{ahmed2016structural} as performance measures. For aerodynamic wing design, researchers have used measures like the lift-to-drag ratio~\cite{chen2019aerodynamic}.

Current state-of-the-art generative models have no mechanism of explicitly promoting design generation with improved performance and diversity. In this work, we focus on addressing the problem of simultaneously maximizing diversity and (possibly multivariate) performance of generated designs. Specifically, we develop a new loss function, based on Determinantal Point Processes (DPPs)~\cite{kulesza2012determinantal}, for generative models to encourage both high-performance and diverse design generation. Using this loss function, we develop a new variant of GAN, named MO-PaDGAN (Multi-Objective Performance Augmented Diverse Generative Adversarial Network). We show that it can generate new samples with a better coverage of the design space and improvement in all performance measures compared to a baseline GAN. More importantly, we found that MO-PaDGAN can expand the existing boundary of the design space towards high-performance regions outside the training data, which indicates its ability of generating novel high-performance designs.

% Our work is based on the recently proposed PaDGAN~\cite{chen2020padgan}, which uses DPP-based loss function to encourage high-performance diverse design generation. However, unlike PaDGAN's goal of scalar performance, we focus on multiple performance measures for designs. 
One closely related work is the GDPP method~\cite{elfeki2019gdpp}, where the authors devised an objective term that matches the diversity of generated data with training data. The diversity is modeled by the DPP kernel.
MO-PaDGAN differs from this method in two aspects. 
% First, MO-PaDGAN is stable against scaling of data while on validating GDPP for multiple test problems, we found that their method does not work for problems with training data at different scales. 
First, MO-PaDGAN aims to maximize the diversity of generated samples rather than matching it with training data. Thus, MO-PaDGAN can generate diverse samples even when the original training data is biased in favor of a few modes, while GDPP will mimic the bias in generated samples. 
Second, GDPP does not consider the performance of generated samples, whereas we incorporate (possibly multivariate) performance measurements into the DPP kernel and encourage generation of high-performance samples. This is important in engineering design settings as we want the generated designs to not only look realistic, but also be useful. The contributions and novelty of this work are as follows:
\setlist{nolistsep}
\begin{enumerate}[noitemsep]
    \item We propose a novel design generation method that simultaneously encourage generation of diverse and high-performance designs.
    \item We propose a way to incorporate multivariate performance measurements into the DPP kernel-based loss function of GAN, so that the generated samples have higher average and peak performance than training data in all dimensions.
    \item We find that MO-PaDGAN can expand the design space boundary towards high-performance regions that it had not seen from existing data.
    %  \item We propose a way to control the trade-off between performance and diversity in DPPs. Our method extends past work on decomposing a DPP kernel by providing a way to tune the relative importance of performance over diversity.
\end{enumerate}

%%%%%%%%%%%%%%%%%%%%%%%%%%%%%%%%%%%%%%%%%%%%%%%%%%%%%%%%%%%%%%%%%%%%%%
\section{Background}

Below we provide background on GANs and DPP kernels.

\subsection{Generative Adversarial Nets}

Generative Adversarial Networks~\cite{goodfellow2014generative} model a game between a generative model (\textit{generator}) and a discriminative model (\textit{discriminator}). The generator $G$ maps an arbitrary noise distribution to the data distribution (\ie, the distribution of designs in our scenario), thus can generate new data; while the discriminator $D$ tries to perform classification, \ie, to distinguish between real and generated data. Both $G$ and $D$ are usually built with deep neural networks. As $D$ improves its classification ability, $G$ also improves its ability to generate data that fools $D$. 
Thus, a GAN has the following objective function:
 \begin{equation}
 \begin{split}
 \min_G\max_D V(D,G) = \mathbb{E}_{\mathbf{x}\sim P_{data}}[\log D(\mathbf{x})] +\\ \mathbb{E}_{\mathbf{z}\sim P_{\mathbf{z}}}[\log(1-D(G(\mathbf{z})))],
 \label{eq:gan_loss}
 \end{split}
 \end{equation}
where $\mathbf{x}$ is sampled from the data distribution $P_{data}$, $\mathbf{z}$ is sampled from the noise distribution $P_{\mathbf{z}}$, and $G(\mathbf{z})$ is the generator distribution. 
A trained generator thus can map from a predefined noise distribution to the distribution of designs. The noise input $\mathbf{z}$ is considered as the latent representation of the data, which can be used for design synthesis and exploration. Note that GANs often suffer from \textit{mode collapse}~\cite{salimans2016improved}, where the generator fails to capture all modes of the data distribution. In this work, by maximizing the diversity objective, mode collapse is discouraged as it leads to less diverse samples.

\subsection{Decomposition of a DPP kernel}

DPP kernels can be decomposed into quality and diversity parts~\cite{kulesza2012determinantal}. 
The $(i,j)^{th}$ entry of a positive semi-definite DPP kernel $L$ can be expressed as:
\begin{equation}
 L_{ij} = q_i\;\phi(i)^T\;\phi(j)\;q_j.
 \label{eq:L_ij}
\end{equation}

We can think of $q_i \in R^+$ as a scalar value measuring the quality of an item $i$, and $\phi(i)^T\;\phi(j)$ as a signed measure of similarity between items $i$ and $j$. The decomposition enforces $L$ to be positive semidefinite. 
%This independent modeling of quality and diversity allows us to combine them into a unified model. 
Suppose we select a subset $S$ of samples, then this decomposition allows us to write the probability of this subset $S$ as the square of the volume spanned by $q_i \phi _i$ for $i \in S$ using the equation below:
\begin{equation}\label{eq:eq4}
    \mathbb{P}_L(S)~\propto~\prod_{i \in S} ({q_i}^2) \det(K_S),
\end{equation}
where $K_S$ is the similarity matrix of $S$. %The first term increases with the quality of the selected items, and the second term increases with the diversity of the selected items. 
As item $i$’s quality $q_i$ increases, so do the probabilities of sets containing item $i$. As two items $i$ and $j$ become more similar, ${\phi_i}^T \phi_j$ increases and the probabilities of sets containing both $i$ and $j$ decrease. The key intuition of MO-PaDGAN is that if we can integrate the probability of set selection from Eq.~(\ref{eq:eq4}) to the loss function of any generative model, then while training it will be encouraged to generate high probability subsets, which will be both diverse and high-performance.

%%%%%%%%%%%%%%%%%%%%%%%%%%%%%%%%%%%%%%%%%%%%%%%%%%%%%%%%%%%%%%%%%%%%%%
\section{Methodology}
\label{sec:methodology}

MO-PaDGAN adds a \textit{performance augmented DPP loss} to a standard GAN architecture which measures the diversity and performance of a batch of generated designs during training. The overall model architecture of MO-PaDGAN is shown in Fig.~\ref{fig:architecture}. 
We describe the DPP kernel part next.
% In this section, we begin by describing how to decompose a DPP kernel, then proceed on how to create a DPP loss which augments high performing designs, and finally provide a method to balance diversity and quality using a quality dial.  
%We explain this new loss and the training method in the following sections.

%%%%%%%%%%%%%%%%%%%%%%%%%%%%%%%%%%%%%%%%%%%%%%%%%%%%%%%%%%%%%%%%%%%%%%
\subsection{Creating a DPP kernel}
We create the kernel $L$ for a sample of points generated by MO-PaDGAN from known inter-sample similarity values and performance vector. 

The similarity terms $\phi(i)^T \phi(j)$ can be derived using any similarity kernel, which we represent using $k(\mathbf{x}_i,\mathbf{x}_j) = \phi(i)^T \phi(j)$ and $\| \phi(i) \| = \| \phi(j) \| = 1$.
Here $\mathbf{x}_i$ is a vector representation of a design. Note that in a DPP model, the quality of an item is a scalar value representing design performance such sa compliance, displacement, drag-coefficient.
We can estimate the performance using an external model (like a physics-based simulator).
% or by finding the distance of current performance of a design from a target performance. 
For multivariate performance, we use a \textit{performance aggregator} to obtain a scalar quality
$q(\mathbf{x})=\sum_{j=1}^K w_j p_j(\mathbf{x})$, where $p_1,...,p_K$ are K-dimensional performances and the corresponding weights $w_1,...,w_K$ are positive numbers sampled uniformly at random and sum to 1. Maximizing $q(\mathbf{x})$ pushes the non-dominated Pareto set of generated samples in the performance space to have higher values. While more complex performance aggregators (\eg, the Chebyshev distance from an ideal performance vector) are also applicable, we used the common linear scalarization to have fewer assumptions about the performance space. %The weighted sum method is commonly used in multi-objective optimization methods.
% By taking multiple random samples of weight vectors for the same design, multiple quality measures can be obtained which capture all performance dimensions.

%%%%%%%%%%%%%%%%%%%%%%%%%%%%%%%%%%%%%%%%%%%%%%%%%%%%%%%%%%%%%%%%%%%%%%
\subsection{Performance Augmented DPP Loss}

Our performance augmented DPP loss models diversity and performance simultaneously and gives a lower loss to sets of designs which are both high-performance and diverse. Specifically, we construct a kernel matrix $L_B$ for a generated batch $B$ based on Eq.~(\ref{eq:L_ij}). For each entry of $L_B$, we have
\begin{equation}
L_B(i,j) = k(\mathbf{x}_i,\mathbf{x}_j)\left(q(\mathbf{x}_i)q(\mathbf{x}_j)\right)^{\gamma_0},
\label{eq:L_B}
\end{equation}
where $\mathbf{x}_i,\mathbf{x}_j \in B$, $q(\mathbf{x})$ is the quality function at $\mathbf{x}$, and $k(\mathbf{x}_i,\mathbf{x}_j)$ is the similarity kernel between $\mathbf{x}_i$ and $\mathbf{x}_j$. 
For a given kernel, DPP decomposition does not allow us to change the trade-off between quality and diversity. To allow this, we adjust the dynamic range of the quality scores by using an exponent ($\gamma_0$) as a parameter to change the distribution of quality. A larger $\gamma_0$ increases the relative importance of quality as compared to diversity, which provides the flexibility to a user of MO-PaDGAN in deciding emphasis on quality vs diversity.

The performance augmented DPP loss is expressed as
\begin{equation}
\pazocal{L}_{\text{PaD}}(G) = -\frac{1}{|B|}\log\det(L_B) = -\frac{1}{|B|}\sum_{i=1}^{|B|} \log\lambda_i,
\label{eq:pad_loss}
\end{equation}
where $\lambda_i$ is the $i$-th eigenvalue of $L_B$. We add this loss to the vanilla GAN's objective in Eq.~(\ref{eq:gan_loss}) and form a new objective:
\begin{equation}
\min_G\max_D V(D,G) + \gamma_1 \pazocal{L}_{\text{PaD}}(G),
\label{eq:overall_loss}
\end{equation}
where $\gamma_1$ controls the weight of $\pazocal{L}_{\text{PaD}}$(G). 
For the backpropogation step, to update the weight $\theta_G^i$ in the generator in terms of $\pazocal{L}_{\text{PaD}}(G)$, we descend its gradient based on the chain rule:
\begin{equation}
\frac{\partial \pazocal{L}_{\text{PaD}}(G)}{\partial \theta_G^i} = \sum_{j=1}^{|B|} \left( \frac{\partial \pazocal{L}_{\text{PaD}}(G)}{\partial q(\mathbf{x}_j)} \frac{d q(\mathbf{x}_j)}{d \mathbf{x}_j} + \frac{\partial \pazocal{L}_{\text{PaD}}(G)}{\partial \mathbf{x}_j} \right) \frac{\partial \mathbf{x}_j}{\partial \theta_G^i},
\label{eq:gradient}
\end{equation}
where $\mathbf{x}_j = G(\mathbf{z}_j)$. Equation~(\ref{eq:gradient}) indicates a need for $dq(\mathbf{x})/d\mathbf{x}$, which is the gradient of the quality function. In practice, this gradient is accessible when the quality is evaluated through a performance estimator that is differentiable, like adjoint-based solver methods. If the gradient of a performance estimator is not available, one can either use numerical differentiation or approximate the quality function using a differentiable surrogate model (\eg, a neural network-based surrogate model, as used in our experiments).

%%%%%%%%%%%%%%%%%%%%%%%%%%%%%%%%%%%%%%%%%%%%%%%%%%%%%%%%%%%%%%%%%%%%%%
\section{Experimental Results}

In this section, we demonstrate the merit of modeling performance and diversity simultaneously by applying MO-PaDGAN on a real-world airfoil shape generation problem and comparing it against a vanilla GAN.

An airfoil is the cross-sectional shape of a wing or a propeller/rotor/turbine blade. In this example, we use the UIUC airfoil database\footnote{\url{http://m-selig.ae.illinois.edu/ads/coord_database.html}} as our data source. It provides the geometries of nearly 1,600 real-world airfoil designs. We preprocessed and augmented the dataset based on \citet{chen2019aerodynamic} to generate a dataset of 38,802 airfoils, each of which is represented by 192 surface points (\ie, $\mathbf{x}_i\in \mathbb{R}^{192\times 2}$). We use two performance measures for designing the airfoils~\textemdash~the lift coefficient ($C_L$) and the lift-to-drag ratio ($C_L/C_D$). These two are common objectives in aerodynamic design optimization problems and have been used in different multi-objective optimization studies~\cite{park2010optimal}. We use XFOIL~\cite{drela1989xfoil} for computational fluid dynamics (CFD) simulations and compute $C_L$ and $C_D$ values\footnote{We set $C_L=C_L/C_D=0$ for unsuccessful simulations.}. We scaled the performance scores between 0 and 1. To provide the gradient of the quality function for Eq.~(\ref{eq:gradient}), we trained a neural network-based surrogate model on all 38,802 airfoils to approximate both $C_L$ and $C_D$.

To demonstrate the effectiveness of MO-PaDGAN, we compare it with a vanilla GAN.
We use a RBF kernel with a bandwidth of 1 when constructing $L_B$ in Eq.~(\ref{eq:L_B}), \ie, $k(\mathbf{x}_i,\mathbf{x}_j)=\exp(-0.5\|\mathbf{x}_i-\mathbf{x}_j\|^2)$. We set $\gamma_0=5$ and $\gamma_1=0.2$ for MO-PaDGAN. We used a residual neural network (ResNet)~\cite{he2016deep} as the surrogate model and a B\'ezierGAN~\cite{chen2019aerodynamic,chen2018bezier} to generate airfoils. For simplicity, we refer to the B\'ezierGAN as a vanilla GAN and the B\'ezierGAN with loss $\pazocal{L}_{PaD}$ as a MO-PaDGAN in the rest of the paper.

\begin{figure}[ht]
\centering
\includegraphics[width=0.5\textwidth]{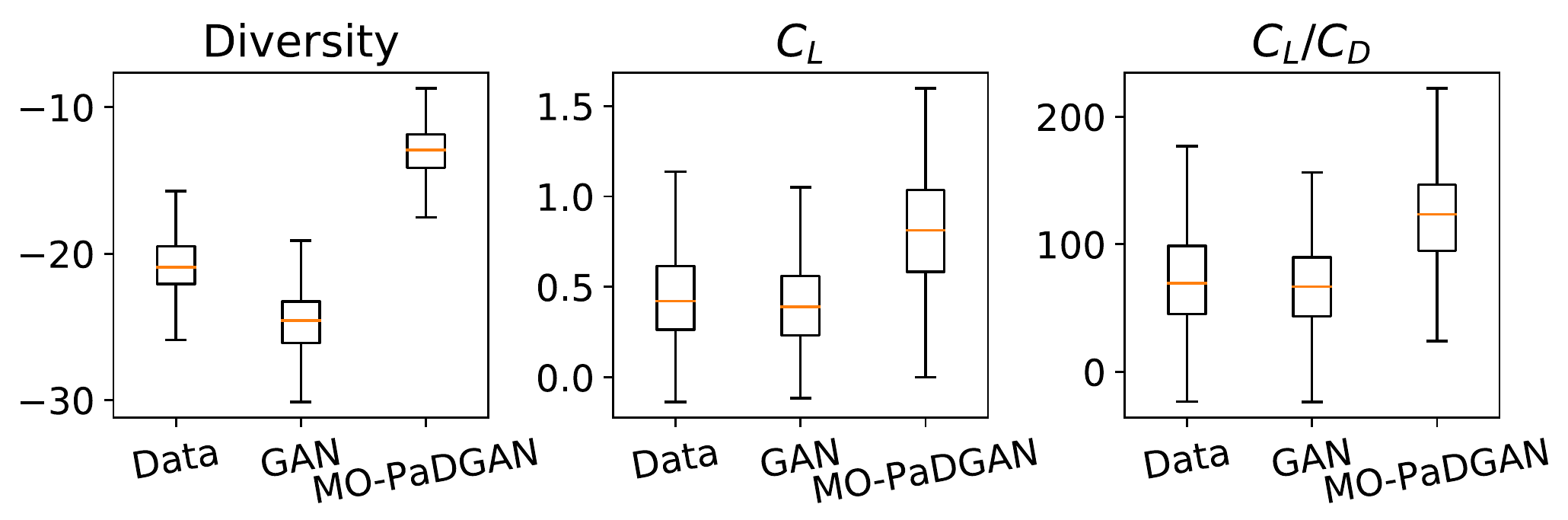}
\vspace*{-8mm}
\caption{Diversity and performance statistics of randomly sampled airfoils.}
\label{fig:airfoil_scores}
\end{figure}

We measure the diversity of generated designs using the log determinant of the similarity matrix:
\begin{equation}
\text{Diversity} = \log\det(L_{S_i}),
\label{eq:div_score}
\end{equation}
where $S_i\subseteq Y$ is a random subset of $Y$ (the set of generated samples or training data), and $L_{S_i}$ is the similarity matrix of $S_i$ with entries $L_{S_i}(j,k)=k(\mathbf{x}_j,\mathbf{x}_k)$ for each $\mathbf{x}_j,\mathbf{x}_k \in S_i$.
We evaluate the diversity for 1000 times. Each time we randomly sample 100 designs from $Y$ (which contains 1000 airfoils). We show the statistics of computed diversity in Fig.~\ref{fig:airfoil_scores}, together with two performance measures ($C_L$ and $C_L/C_D$) of $Y$. It shows that MO-PaDGAN can generate samples with higher diversity and performances than training data and samples from the vanilla GAN.

\begin{figure}[ht]
\centering
\includegraphics[width=0.45\textwidth]{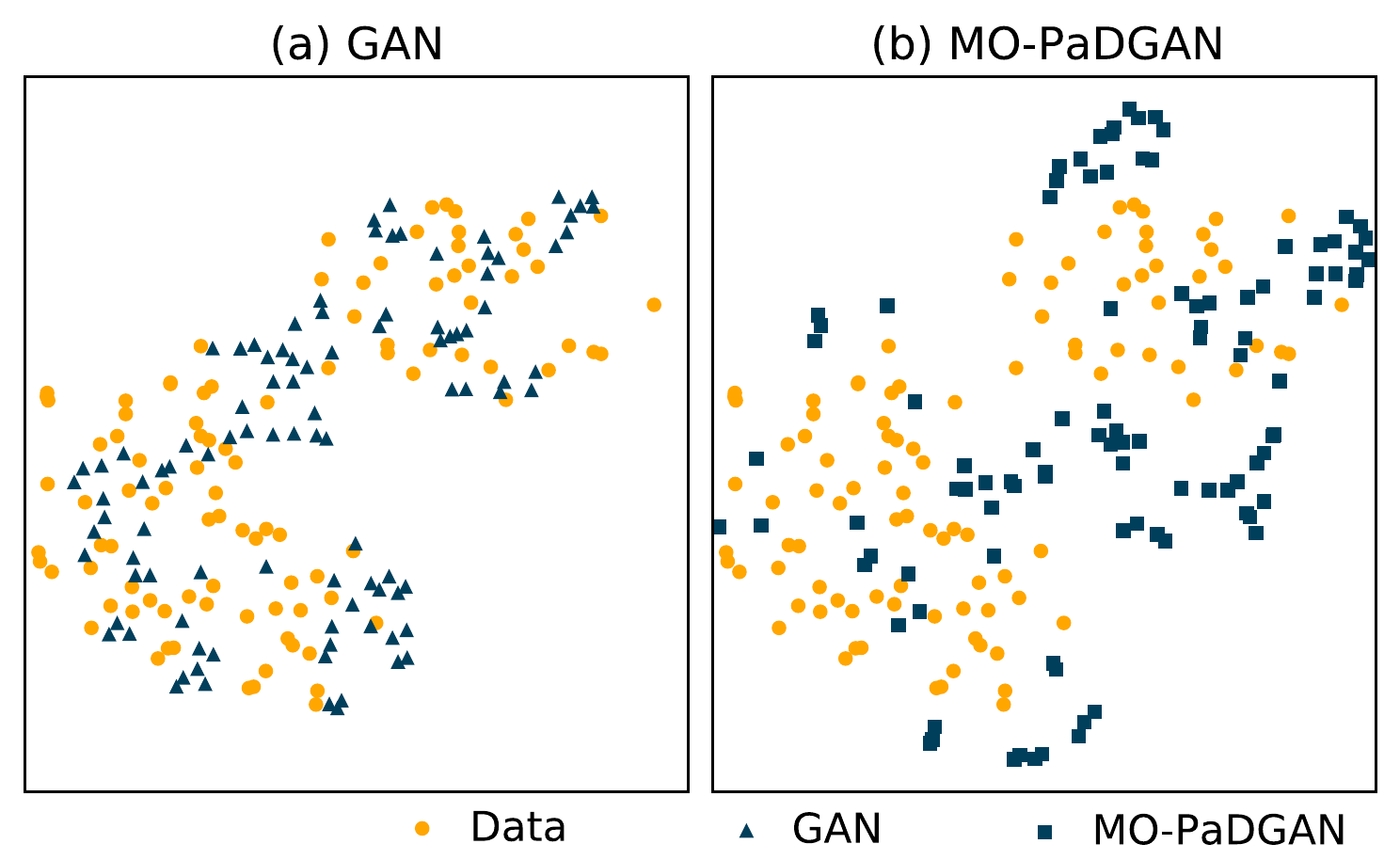}
\vspace*{-3mm}
\caption{Randomly sampled airfoils embedded into a 2D space via t-SNE. MO-PaDGAN expands the boundary of training data.}
\label{fig:airfoils_tsne}
\end{figure}

\begin{figure}[ht]
\centering
\includegraphics[width=0.46\textwidth]{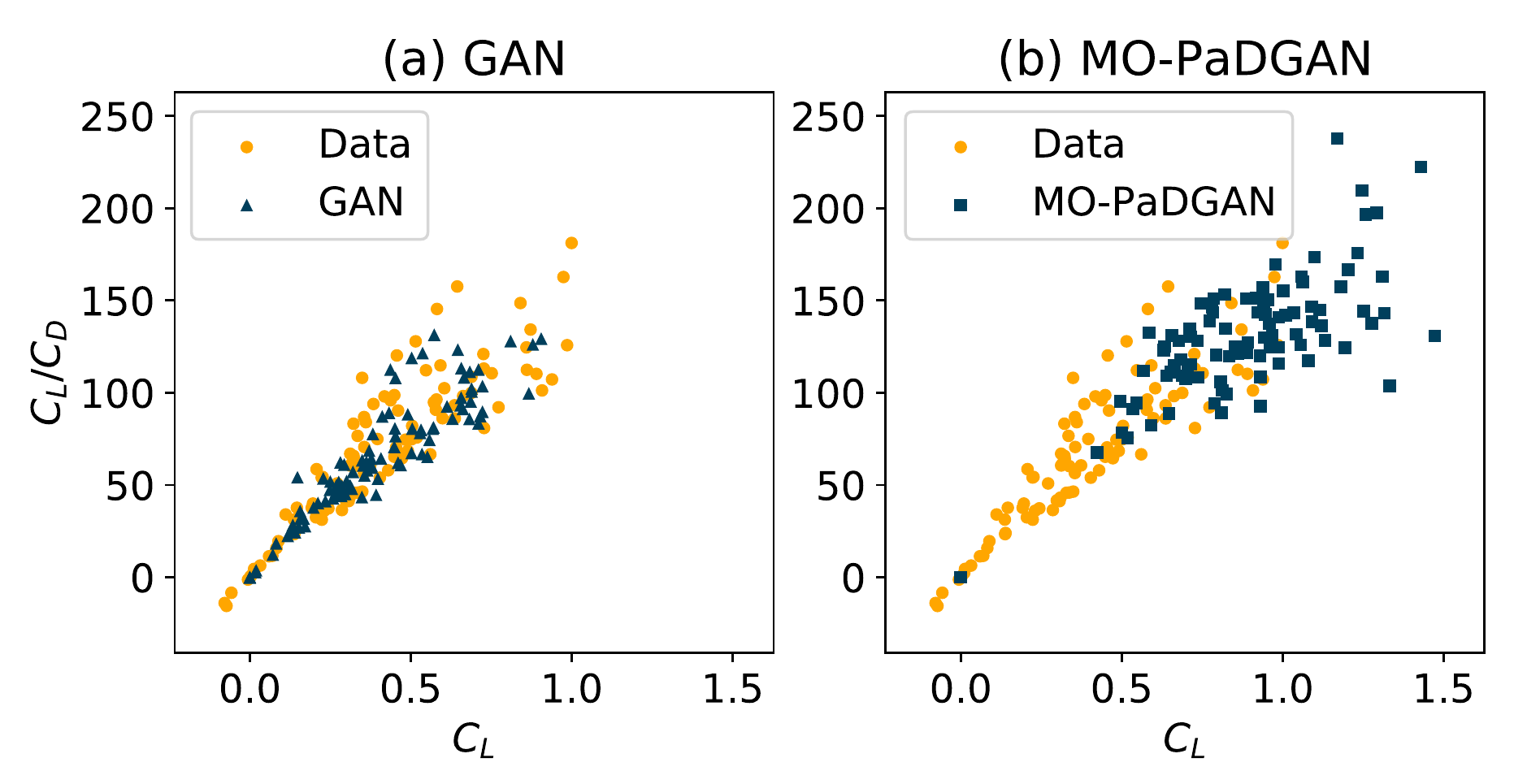}
 \vspace*{-5mm}
\caption{Performance space visualization for airfoils shown in Fig.~\ref{fig:airfoils_tsne} shows MO-PaDGAN improves both performance objectives.}
\label{fig:airfoils_perf}
\end{figure}

To compare the distribution of real and generated airfoils in the design space, we map randomly sampled airfoils into a two-dimensional space through t-SNE, as shown in Figure~\ref{fig:airfoils_tsne}. The results indicate that comparing with a vanilla GAN, MO-PaDGAN generates airfoils that are further away from training data, driven by the DPP loss.

\begin{figure}[ht]
\centering
\includegraphics[width=0.5\textwidth]{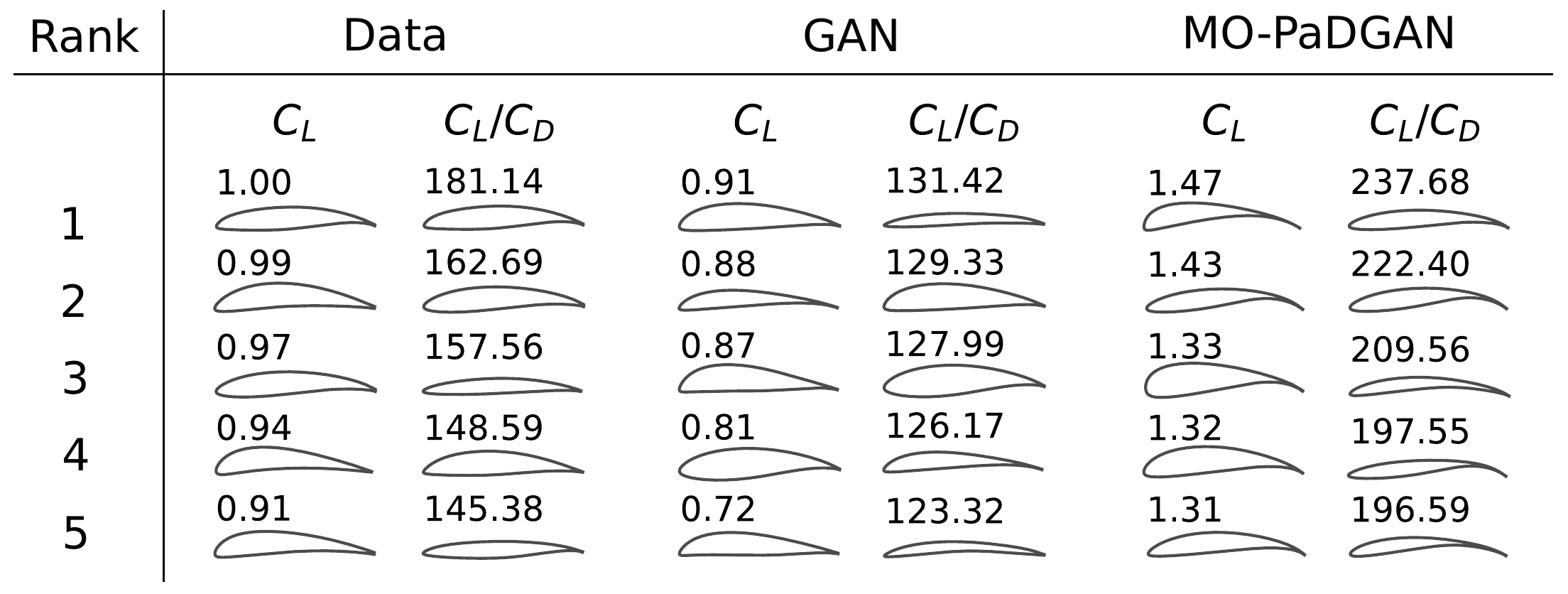}
 \vspace*{-8mm}
\caption{Top five performances and shapes among airfoils shown in Fig.~\ref{fig:airfoils_perf}. We see MO-PaDGAN samples have significantly higher performance than GAN.}
\label{fig:top_airfoils}
\end{figure}

Figure~\ref{fig:airfoils_perf} visualizes the joint distribution of $C_L$ and $C_L/C_D$ for randomly sampled airfoils. It shows that MO-PaDGAN generates airfoils with performances exceed randomly sampled airfoils from training data and the vanilla GAN (\ie, the non-dominated Pareto set of generated samples is pushed further in the performance space to have higher values). Figures~\ref{fig:airfoils_tsne} and \ref{fig:airfoils_perf} indicate that MO-PaDGAN can expand the existing boundary of the design space towards high-performance regions outside the training data. This directed expansion is allowed since we provide the quality gradients (\ie, $dq(\mathbf{x})/d\mathbf{x}$) information to MO-PaDGAN. 
Figure~\ref{fig:top_airfoils} further demonstrates that the top airfoils generated by MO-PaDGAN have much higher performances than those from data and the vanilla GAN (\ie, the performances of the top five airfoils generated by MO-PaDGAN dominates those from training data and the vanilla GAN).

%%%%%%%%%%%%%%%%%%%%%%%%%%%%%%%%%%%%%%%%%%%%%%%%%%%%%%%%%%%%%%%%%%%%%
\section{Conclusion}

We proposed MO-PaDGAN with a new loss function based on Determinantal Point Processes. This model is useful when we want to explore different high-performance design alternatives or discover novel solutions. For example, when performing design optimization, one may accelerate the search for global optimal solutions by sampling start points from the proposed model. It can also be a tool in the early conceptual design stage to aid the creative process. The proposed framework also generalizes to other generative models like VAEs and can be used for various synthesis problems like 3D shape generation and molecule discovery.

\bibliography{asme2e}

\begin{thebibliography}{14}
\providecommand{\natexlab}[1]{#1}
\providecommand{\url}[1]{\texttt{#1}}
\expandafter\ifx\csname urlstyle\endcsname\relax
  \providecommand{\doi}[1]{doi: #1}\else
  \providecommand{\doi}{doi: \begingroup \urlstyle{rm}\Url}\fi

\bibitem[Ahmed et~al.(2016)Ahmed, Deb, and Bhattacharya]{ahmed2016structural}
Ahmed, F., Deb, K., and Bhattacharya, B.
\newblock Structural topology optimization using multi-objective genetic
  algorithm with constructive solid geometry representation.
\newblock \emph{Applied Soft Computing}, 39:\penalty0 240--250, 2016.

\bibitem[Bendsoe \& Sigmund(2004)Bendsoe and Sigmund]{bendsoe_topology_2004}
Bendsoe, M.~P. and Sigmund, O.
\newblock \emph{Topology Optimization: Theory, Methods and Applications}.
\newblock Springer, February 2004.
\newblock ISBN 9783540429920.

\bibitem[Chen \& Fuge(2018)Chen and Fuge]{chen2018bezier}
Chen, W. and Fuge, M.
\newblock B\'eziergan: Automatic generation of smooth curves from interpretable
  low-dimensional parameters.
\newblock \emph{arXiv preprint arXiv:1808.08871}, 2018.

\bibitem[Chen \& Fuge(2019)Chen and Fuge]{chen2019synthesizing}
Chen, W. and Fuge, M.
\newblock Synthesizing designs with interpart dependencies using hierarchical
  generative adversarial networks.
\newblock \emph{Journal of Mechanical Design}, 141\penalty0 (11), 2019.

\bibitem[Chen et~al.(2017)Chen, Fuge, and Chazan]{chen2017design}
Chen, W., Fuge, M., and Chazan, J.
\newblock Design manifolds capture the intrinsic complexity and dimension of
  design spaces.
\newblock \emph{Journal of Mechanical Design}, 139\penalty0 (5), 2017.

\bibitem[Chen et~al.(2019)Chen, Chiu, and Fuge]{chen2019aerodynamic}
Chen, W., Chiu, K., and Fuge, M.
\newblock Aerodynamic design optimization and shape exploration using
  generative adversarial networks.
\newblock In \emph{AIAA SciTech Forum}, San Diego, USA, Jan 2019. AIAA.

\bibitem[Drela(1989)]{drela1989xfoil}
Drela, M.
\newblock Xfoil: An analysis and design system for low reynolds number
  airfoils.
\newblock In \emph{Low Reynolds number aerodynamics}, pp.\  1--12. Springer,
  1989.

\bibitem[Elfeki et~al.(2019)Elfeki, Couprie, Riviere, and
  Elhoseiny]{elfeki2019gdpp}
Elfeki, M., Couprie, C., Riviere, M., and Elhoseiny, M.
\newblock Gdpp: Learning diverse generations using determinantal point
  processes.
\newblock In \emph{International Conference on Machine Learning}, pp.\
  1774--1783, 2019.

\bibitem[Goodfellow et~al.(2014)Goodfellow, Pouget-Abadie, Mirza, Xu,
  Warde-Farley, Ozair, Courville, and Bengio]{goodfellow2014generative}
Goodfellow, I., Pouget-Abadie, J., Mirza, M., Xu, B., Warde-Farley, D., Ozair,
  S., Courville, A., and Bengio, Y.
\newblock Generative adversarial nets.
\newblock In \emph{Advances in neural information processing systems}, pp.\
  2672--2680, 2014.

\bibitem[He et~al.(2016)He, Zhang, Ren, and Sun]{he2016deep}
He, K., Zhang, X., Ren, S., and Sun, J.
\newblock Deep residual learning for image recognition.
\newblock In \emph{Proceedings of the IEEE conference on computer vision and
  pattern recognition}, pp.\  770--778, 2016.

\bibitem[Kingma \& Welling(2013)Kingma and Welling]{kingma2013auto}
Kingma, D.~P. and Welling, M.
\newblock Auto-encoding variational bayes.
\newblock \emph{arXiv preprint arXiv:1312.6114}, 2013.

\bibitem[Kulesza \& Taskar(2012)Kulesza and Taskar]{kulesza2012determinantal}
Kulesza, A. and Taskar, B.
\newblock Determinantal point processes for machine learning.
\newblock \emph{arXiv preprint arXiv:1207.6083}, 2012.

\bibitem[Park \& Lee(2010)Park and Lee]{park2010optimal}
Park, K. and Lee, J.
\newblock Optimal design of two-dimensional wings in ground effect using
  multi-objective genetic algorithm.
\newblock \emph{Ocean Engineering}, 37\penalty0 (10):\penalty0 902--912, 2010.

\bibitem[Salimans et~al.(2016)Salimans, Goodfellow, Zaremba, Cheung, Radford,
  and Chen]{salimans2016improved}
Salimans, T., Goodfellow, I., Zaremba, W., Cheung, V., Radford, A., and Chen,
  X.
\newblock Improved techniques for training gans.
\newblock In \emph{Advances in neural information processing systems}, pp.\
  2234--2242, 2016.

\end{thebibliography}
\bibliographystyle{icml2020}

% \appendix
% \section{Do \emph{not} have an appendix here}

% \textbf{\emph{Do not put content after the references.}}
% %
% Put anything that you might normally include after the references in a separate
% supplementary file.

% We recommend that you build supplementary material in a separate document.
% If you must create one PDF and cut it up, please be careful to use a tool that
% doesn't alter the margins, and that doesn't aggressively rewrite the PDF file.
% pdftk usually works fine. 

% \textbf{Please do not use Apple's preview to cut off supplementary material.} In
% previous years it has altered margins, and created headaches at the camera-ready
% stage. 
% %%%%%%%%%%%%%%%%%%%%%%%%%%%%%%%%%%%%%%%%%%%%%%%%%%%%%%%%%%%%%%%%%%%%%%%%%%%%%%%
%%%%%%%%%%%%%%%%%%%%%%%%%%%%%%%%%%%%%%%%%%%%%%%%%%%%%%%%%%%%%%%%%%%%%%%%%%%%%%%

\end{document}